# The Cyborg Astrobiologist: Matching of Prior Textures by Image Compression for Geological Mapping and Novelty Detection

P.C. McGuire (*,1, *formerly at:* 2), A. Bonnici (3), K.R. Bruner (4), C. Gross (1),

J. Ormö (5), R.A. Smosna (4), S. Walter (1), L. Wendt (1);

(1) Freie Universität, Berlin, Germany,

(2) University of Chicago, Chicago, IL, USA,

(3) University of Malta, Malta,

(4) West Virginia University, Morgantown, WV, USA,

(5) Centro de Astrobiología, CSIC-INTA, Torrejón de Ardoz, Madrid, Spain,

(*, mcguirepatr@gmail.com)

## Abstract

We describe an image-comparison technique of Heidemann and Ritter (2008a,b) that uses image compression, and is capable of: (i) detecting novel textures in a series of images, as well as of: (ii) alerting the user to the similarity of a new image to a previously-observed texture. This image-comparison technique has been implemented and tested using our Astrobiology Phone-cam system, which employs Bluetooth communication to send images to a local laptop server in the field for the image-compression analysis. We tested the system in a field site displaying a heterogeneous suite of sandstones, limestones, mudstones and coalbeds. Some of the rocks are partly covered with lichen. The image-matching procedure of this system performed very well with data obtained through our field test, grouping all images of yellow lichens together and grouping all images of a coal bed together, and giving a 91% accuracy for similarity detection. Such similarity detection could be employed to make maps of different geological units. The novelty-detection performance of our system was also rather good (a 64% accuracy). Such novelty detection may become valuable in searching for new geological units, which could be of astrobiological interest. The current system is not directly intended for mapping and novelty detection of a second field site based upon image-compression analysis of an image database from a first field site, though our current system could be further developed towards this end. Furthermore, the image-comparison technique is an unsupervised technique that is not capable of directly classifying an image as containing a particular geological feature; labeling of such geological features is done *post facto* by human geologists associated with this study, for the purpose of analyzing the system's performance. By providing more advanced capabilities for similarity detection and novelty detection, this image-compression technique could be useful in giving more scientific autonomy to robotic planetary rovers, and in assisting human astronauts in their geological exploration and assessment.

# 1. Introduction

In prior work, we have developed computer algorithms for real-time novelty detection and rarity mapping for astrobiological and geological exploration (Bartolo *et al.* 2007; Gross *et al.* 2009, 2010; McGuire *et al.* 2004, 2005a,b, 2010; Wendt *et al.* 2009). These algorithms were tested at astrobiological field sites using mobile computing platforms – originally (McGuire *et al.* 2004, 2005a,b, 2010) with a wearable computer connected to a digital video camera, but more recently (Bartolo *et al.* 2007; Gross *et al.* 2009, 2010; McGuire *et al.* 2010; Wendt *et al.* 2009) with a phone camera connected wirelessly to a local or remote server computer. The image features used in the novelty detection and rarity mapping in prior work were based only upon RGB or HSI color.

Nonetheless, even with image features limited to color, the mobile exploration systems worked very robustly. Based on an analysis of the procedures governing the team´s geologists when analyzing an encountered outcrop, we developed a concept of "novelty detection" for guiding the Cyborg Astrobiologist system in a first step towards mimicking the geologist´s approach. Very simplified, the geologist´s decisions when approaching an outcrop are based upon observations that may be useful to "tell the tale" about the outcrop, i.e. relations between individual parts of the outcrop. This can, for instance, be contacts between different lithologies, alteration sequences, diagenetic variations, or structures such as beddings or fractures. Commonly, such variations are, at a first distant glance of an outcrop, expressed as color and/or texture variations. In, for instance, the case of color variations, the geologist may then decide that it is a certain color that stands out from a more common background that should be given special attention, i.e. as "interest points". The next step for the geologists is, thus, usually to investigate the contacts between these first-order "interest points" and the background. When further advancing on the outcrop, new "interest points" of more detailed scale are then selected based upon the same principle in a somewhat iterative way. The 'color-only' Cyborg Astrobiologist has in our first tests been employed in unvegetated, desert-like environments where it was able to identify novel or uncommon areas in image sequences, ranging from mostly white-colored gypsum to mostly red-colored 'red bed' sandstones. The system was able to identify, for example, bleached areas and hematite concretions in the red beds of Spain, and lichens of varying colors within the desert landscapes of Spain and Utah as being novel features (when first observed) of those landscapes (Gross *et al.* 2009, 2010; McGuire *et al.* 2010; Wendt *et al.* 2009). However, another important factor in addition to color variations in the decision-making of the geologist when approaching an outcrop is texture, the objective of this study.

Herein, using data acquired by the Astrobiology Phone-cam at a geological field site (Fig. 1), we test an image-comparison technique of Heidemann and Ritter (2008a,b) that uses image compression and is capable of (i) detecting novel (colored) textures in a series of images as well as of (ii) alerting the user to the similarity of a new image to a previously-observed texture. We first implemented this technique in 2010 (Bonnici *et al.* 2010), but we did not test this software at a geological field site until now. Such a capability could be useful in giving more scientific autonomy to robotic planetary rovers (Volpe 2003, Fink *et al.* 2005, Castano *et al.* 2007, Halatci, Brooks and Iagnemma 2007, 2008), and perhaps in assisting human astronauts in their geological exploration. For example, suppose a semi-autonomous planetary rover equipped with texture-based novelty detection is observing a long series of textures corresponding to hematite concretions embedded in mineral deposits. With texture-based novelty detection (Thompson *et al.* 2012; Gulick *et al.* 2001, 2004), this rover would report, during stage 1 of its approach towards the outcrop, that a particular previously-unobserved horizontally-layered texture is novel, and hence merits further investigation. On stage 2 it would detect smaller spherical objects, possibly even of different color relative to the background, within the previously observed layers. Depending on the resolution of the camera this process could continue, iteratively, with further observations of onion-layering of the spherules and so forth. A certain amount of autonomous decision-making of the rover would in this way greatly reduce the amount of data to be transferred between the rover and the scientists.

## 2. Heidemann and Ritter's Image-Compression Technique

Following Heidemann and Ritter (2008a,b), we "calculate the similarity of two images $I_1$, $I_2$ as:

$$D_{SIM}(I_1, I_2) = S(I_1) + S(I_2) - S(I_{12}) \,, \quad (1)$$

where $S(\cdot)$ denotes the [(single-number) byte] size of a compressed image. $I_{12}$ is the 'joint' image obtained as juxtaposition of pixel arrays $I_1$ and $I_2$". If the two images are very similar, then $S(I_{12})$ will be small and $D_{SIM}(I_1, I_2)$ will be large. If the two images are very different (due to textural and color differences), then $S(I_{12})$ will be large and $D_{SIM}(I_1, I_2)$ will be small.

The juxtaposition in our implementation is left-right, but we have not compared our implementation to an up-down juxtaposition. Various image-compression programs were investigated by Heidemann and Ritter (2008a,b), and their optimal image compressor for texture classification was *gzip*, which we have chosen as our image compressor[1]. *Gzip* relies on Huffman entropy coding (Huffman, 1952; see also: Saravanan and Ponalagusamy, 2010) and the Lempel-Ziv algorithm (Lempel and Ziv, 1977; see also: Kärkkäinen, Kempa, and Puglisi, 2013). Huffman coding is based on how frequent the symbols[2] are in a data stream, assigning shorter bit-length representations for more-commonly-used symbols. The Lempel-Ziv algorithm (LZ77) eliminates duplicate strings of symbols using 'sliding-window compression' (referring to the buffer window that records the previously-observed symbols in the data stream)[3]. Quoting Heidemann and Ritter (2008b): "The Lempel-Ziv algorithm (LZ77) (Lempel and Ziv, 1977), which detects repeatedly occurring symbol sequences within the data, such that a dictionary can be established. A repeated symbol sequence can then be replaced by the symbol defined in the dictionary."[4] The *gzip* algorithm outperforms (Heidemann and Ritter, 2008b) the correlation technique and the histogram-based matching technique for the texture-classification, object-recognition and image-retrieval tasks.

By employing a simple image-compression technique such as Heidemann and Ritter's, we can avoid the complexities and ambiguities of more-reductionist textural-comparison techniques (i.e., Haralick *et al.* 1973, Rao 2012), wherein a significant number of different textural indicators (at different spatial scales and orientations) are computed and mapped for each image.

---

[1] In our implementation, we have used the *gzip* functionality of *MATLAB* 7.10.0 (R2010a), using the *MATLAB* function java.util.zip.GZIPOutputStream( ), which calls a *Java* function that was derived from the *Java* Deflater compression algorithm. Our version of *Java* is 7.0.110.21. The *Java* Deflater algorithm is based upon the Compressed Data Format specification for DEFLATE in the Request for Comments #1951 (http://www.ietf.org/rfc/rfc1951.txt ) of the Internet Engineering Task Force and the Internet Society.

[2] A symbol in this context is a numeric byte.

[3] We use *gzip*'s standard sliding window size of 32kB. In future work, by modifying the *gzip* source code, we plan to study how texture-matching performance depends on this sliding window size.

[4] Bytes of similar numeric value will not be automatically classified as a single symbol. Therefore, digital noise or slight, intrinsic variations of pixel values may affect the compression results. This is one shortcoming of this dictionary-based technique for data compression. In the future, as suggested by Heidemann and Ritter (2008a,b), other data-compression techniques could be used (or developed) that can handle better the inherent variability of RGB pixel values. Such techniques would be in principle lossy, though lossy compression in the JPEG implementation does not work well, due to the discrete cosine transform that was utilized by JPEG (Heidemann and Ritter, 2008a).

In order to simplify interpretation, we have taken the logarithm of $D_{SIM}$ and added a normalization factor, so that the maximal range of $D_{SIM}$ is 0%-100% for the current database of images. We accomplish this by comparing the newest image to itself and setting the value obtained as the 100% similarity value. One would expect this value of $D_{sim}(I_N, I_N)$ to be slightly smaller than $S(I_N)$ since the two images being compared are the same, but the changes at the boundary (between the 2 identical juxtaposed images) introduces a discrepancy in the value of $D_{sim}(I_N, I_N)$.

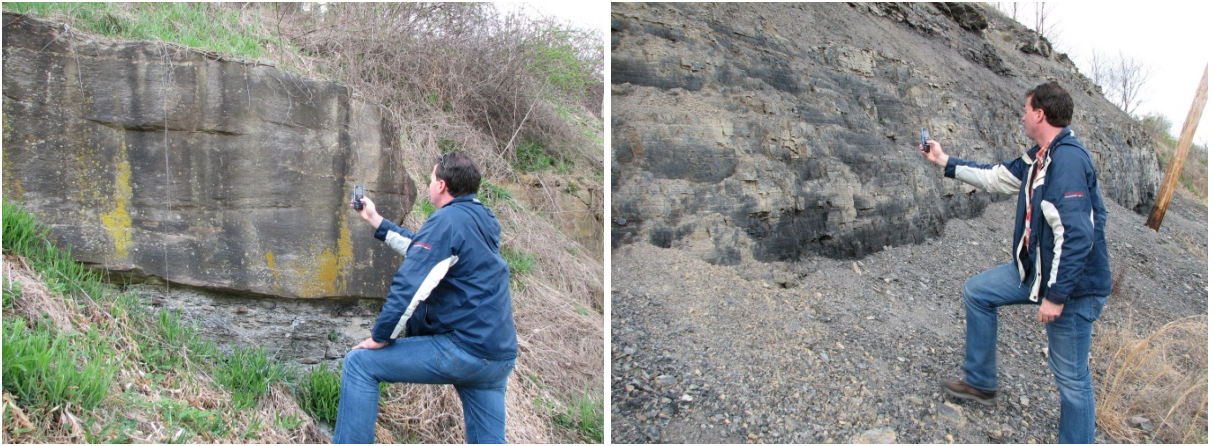

**Figure 1:** The Cyborg Astrobiologist using its image-comparison software with the Astrobiology Phone-cam to study lichens (left picture) and coal with cleats (right picture) at the geological field site (photos by K.R. Bruner).

## 3. Field Tests

The Astrobiology Phone-cam system sends images wirelessly by Bluetooth to a nearby laptop computer, dynamically building an image libary with examples of different terrain. The Phone-cam for this field-test is a Samsung SGH-A767 with 1280×960 RGB color images, and the laptop is a Dell Inspiron 9300. Each incoming image $I_N$ is processed by compiled MATLAB code by the laptop computer by using Eq. 1 to compute the similarity $D_{SIM}(I_N, I_J)$, with each previous image, $I_J$. If $D_{SIM}(I_N, I_J)$ is less than a chosen threshold for all previous images $I_J$ (for all J<N), then the computer considers image $I_N$ as 'novel' and returns a text message to the phone-cam by Bluetooth informing the explorer that the image is novel. If the image is novel, the explorer might decide to perform a more detailed analysis of the ground or rocky outcrop that image $I_N$ represents. If $D_{SIM}(I_N, I_J)$ is greater than the chosen threshold for one or more of the previous images $I_J$, then the image $I_K$, which has the highest similarity score (highest value of $D_{SIM}(I_N, I_K)$), is returned to the phone camera via Bluetooth, juxtaposed with $I_N$, in order for the user to assess the similarity visually. We have performed tests of this procedure for detecting novelty or similarity – a set of example images in the test sequence and their best-matching prior images are juxtaposed in Fig. 2. Image pairs such as in Fig. 2 are the visual information that the image-compression algorithm on the laptop computer outputs and then sends to the phone camera for inspection.

Due to its proximity to three of the authors (McGuire, Smosna and Bruner) at the time of the survey, we chose a geological field site near the northern end of the Morgantown Mall in Morgantown, West Virginia, USA. This mall is the former site of a coal mine, so there are several exposed geological cuts, including one of a coal bed. These rocks[5] comprise a part of West Virginia's coal measures, in particular the lowest section of the Pittsburgh Formation, and have a late Carboniferous age. The outcrop exposes a heterogeneous mix of coal, sandstone, shale, mudstone, and limestone, typically occurring in beds of 30-150 cm thickness. The photographed coal is the Pittsburgh coal, about 3 m thick with well-developed cleats (fractures) and elemental sulfur on the face. The sandstone is thin-bedded to massive with planar bedding and cross-bedding. The shale is characteristically thinly laminated. The mudstone shows a nodular character and locally contains ironstone concretions (composed of iron-carbonate and iron-silicate minerals). The limestone exhibits spherical weathering and solution features.

---

[5] unpublished data from the West Virginia Geological & Economic Survey.

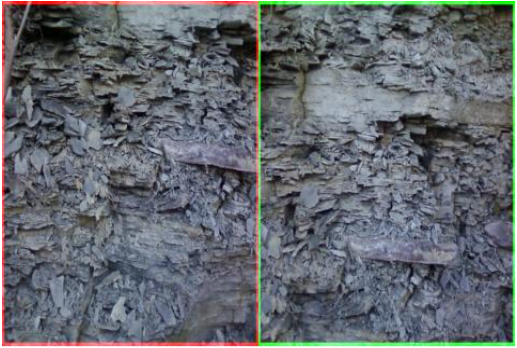

**A) Platy rock-texture (laminated shale)**

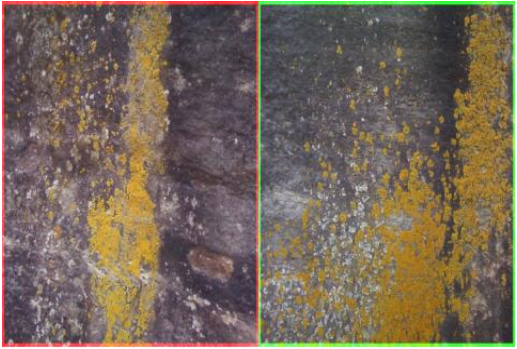

**B) Yellow sporing-bodies of Lichens**

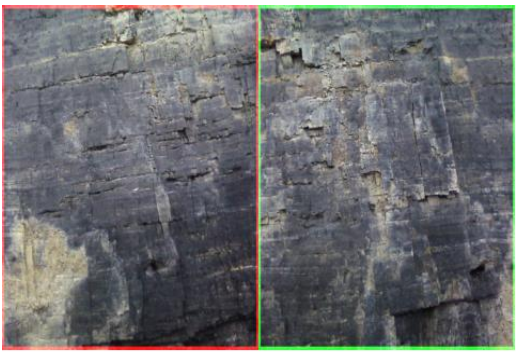

**C) Coal Bed**

**Figure 2**: Good matches: Incoming image $I_N$ is on the left, and best-matching image is on the right. Each image is about 0.5 meters across.

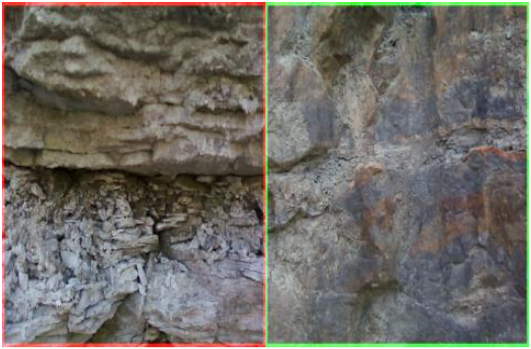

A) **Novel platy rock-texture**

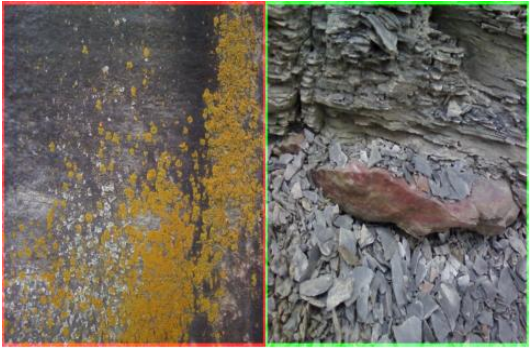

B) **Novel Yellow sporing-bodies of Lichens**

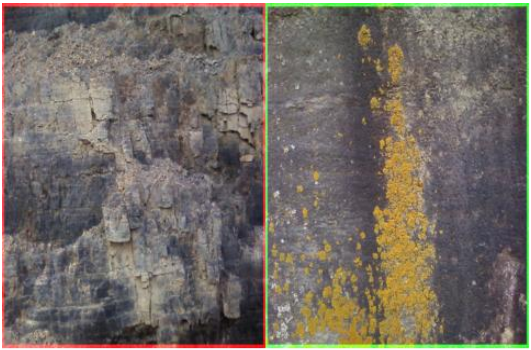

C) **Novel Coal Bed**

**Figure 3**: Novel images: Incoming image $I_N$ is on the left, and best-matching image is on the right. Each image is about 0.5 meters across.

Our original plan was to use the Cyborg Astrobiologist's image-compression system at the geological field site, acquiring and analyzing about 25 images in the 1.5 hour battery lifetime of the laptop server computer (it takes an average of 3-4 minutes to analyze each image). However, at the field site, we decided to acquire 55 images[6] with the Astrobiology Phone-cam in 1 hour, and then to analyze the images offline after the field work was completed. This offline analysis[7] lasted for about 4 hours, longer than the battery life of the laptop. Therefore, this was not the ideal real-time test, but all the capabilities of the system were otherwise tested and more images could be analyzed than during a real-time test.

Three of the 55 result images are shown in Fig. 2. Fig. 2A shows the platy-rock texture on the left, and the best-matching of the prior images on the right. This is indeed a successful match. Likewise, it was a successful match for the yellow-sporing bodies of the lichens in Fig. 2B, and for the coal-bed in Fig. 2C. Generally, for each image which was not novel, the computer successfully matched it best with another image of similar texture and/or color-statistics.

For those images which were novel, the computer either gave it a low $D_{SIM}$ score (<40%) or it matched it with a higher score with an image which even the human analyst would say is similar. See Fig.3 for examples of the novel images and their best matches in the database. Fig. 3A shows the low $D_{SIM}$ score for the novel platy texture with the best-match smooth iron-stained texture. Fig. 3B shows the low $D_{SIM}$ score for the novel yellow sporing bodies of the lichens with the best-match platy-texture with red clast. And Fig. 3C shows the higher $D_{SIM}$ score for the novel coal-bed with the best-match yellow sporing bodies of the lichens. In Fig. 3C, the score is likely higher because the substrate rock for the yellow lichen is black, much like a large part of the coal bed, and because the coal bed has some pale yellow coloring due to sulfur content.

---

[6] The images were acquired without horizon, or rock-debris aprons, or vegetation in the view. This is not the way a robot or rover would have taken the images, unless it was trained to do so. However, it allows us to focus our investigation on the rocky outcrops. We also tried to focus as much as possible on the 'end-member homogeneous textures', as opposed to images that contain multiple, different, mixed textures. In a limited investigation which we do not present here, our software seems to work also with mixed colored textures, wherein the best matching images are of similar mixing amounts of the different colored textures. Much more work needs to be done for these mixed textures.

[7] The testing procedure consisted of (i) an empty image database at the beginning of the test, and subsequently, (ii) each image was compared to the previous ones as it is added to the database, and (iii) to test the performance of the algorithm, the images were later classified based on expert knowledge as one of the following classes: weathered rock, laminar shale… (and so on). The match was considered successful if the algorithm returned an image from the same rock unit.

In Table 1, we estimate the number of images that are novel, similar, or different, as judged by three members of our human team[8], and the number of images that are novel but have high similarity scores, as well as the number of images which are similar and the number of images which are similar with low similarity scores. A more detailed compilation of our results is shown in Table 2, listing for each of the 55 incoming images: (i) which prior image matched best, as well as (ii) notes about novelty and mismatching. The range of observed scores for matching images is generally about 40-50%. Lower scores generally signify novelty, though there are some novel images with scores above 40%, due to similarity to other geological units (see Table 1). The threshold of 40% between similarity and novelty was chosen after the experiment, and it is not a perfect threshold, but it is approximately the right value for these images.

There are examples where the ~48% score appears rather low (for instance, the pair 273/272, where the two images appear to the human observer identical, apart from a small vertical shift). We would expect that the algorithm should be able to detect this better: if the strategy is to compare horizontal scan line (pieces), then the vertical shift should leave many scan line similarities intact and a high value should result (but there might still be high sensitivity to rotation). Based upon prior tests, we believe that the reason the highest scores are not near 100% is because of the discontinuity between the juxtaposed images.

## 4. Discussion

The intent of our algorithm is for analyzing a series of images made in a homogenous environment (a common outcrop) which slightly changes over the time. The scope here is to automatically detect sudden variations in the outcrop's visual appearance, for the purpose of minimizing interactions by the human scientists who are supervising the robotic rover from afar. We do not expect our current algorithm to be able to extrapolate to a new field site in order to identify lichens or coal or other colored textures that were observed at a prior field site, for the following reasons:

---

[8] Novelty is based upon the history of images, and difference is only for the non-novel and non-similar images and is based upon only the labels of the best matching pair.

First, these colored textures likely would have significant intrinsic variability in their color or textural qualities from one site to the next. However, if the color and textural qualities were the same or similar, then the similarity matching and novelty detection would work well. Second, we have not controlled for lighting conditions or photometric angles of observation. We do not attempt such control in the present work, since our main objective is to test the similarity-matching and novelty-detection capabilities of the image comparison algorithm. Nonetheless, in a much-more-advanced system, real-time photometric and atmospheric correction could be added, similar to the post-processing correction system for the CRISM camera on the Mars Reconnaissance Orbiter (McGuire *et al.* 2008). We did not need such photometric or atmospheric control for the field tests completed here, since the lighting conditions were the same (overcast) throughout the one-hour field test.

Third, at the new field site, the color and textural properties of the images may have different spatial scales than at the original field site. Careful control of the camera distance to the rock outcrop may lessen the differences in spatial scale of the colored textures in the images from field site to field site. In the current field tests, we have only tested the system at a single image scale or resolution, given by the human operator locating the camera about 1 meter away from the outcrop. This single-scale property of the images could be approximated on a Mars rover for example by using an imaging camera held by the robotic arm at a fixed distance from the outcrop. We do not have plans currently to develop an algorithm that is capable of distinguishing between colored textures at different spatial scales.

Fourth, this algorithm is an unsupervised algorithm; it does not need to know prior image classifications in order to identify novel or familiar colored-textures. The algorithm does not know that an image is an image of a lichen or a coal-bed – it just can classify an image as familiar or novel, based upon similarity to previous images. So, the algorithm cannot really identify lichens or coal as being lichens or coal. The algorithm does not need to know what it is looking at ahead of time. The classifications were made by human geologists a couple of days after the field test. The intent of the algorithm is to allow the robotic astrobiologist to find those areas of a rocky outcrop that are similar to each other (for mapping purposes) and those areas of a rocky outcrop that are novel (for further investigation and/or sample acquisition).

These conditions exactly correspond to the requirements for an extraterrestrial robotic image-acquisition system, where many images are autonomously taken in a mostly homogeneous environment. Sudden changes or exceptions in the environment could be flagged and reported to the human scientists on Earth, effectively scanning the immense number of images for anomalies (possibly of astrobiological interest).

**Table 1**

|  |  | Truth (human judgement) | | | Sum |
|---|---|---|---|---|---|
|  |  | Novel | Similar | Different |  |
| Predicted by Cyborg Astrobiologist | Novel or Different (Score <40%) | 9 | 3 | 2 | 14 |
|  | Similar (Score ≥40%) | 5 | 31 | 4 | 40 |
| Sum |  | 14 | 34 | 6 | 54 |

Accuracy of Novelty detection: 9/14 =64%

Accuracy of Similarity detection: 31/34 =91%

Accuracy of Difference detection: 2/6=33%

Overall accuracy: 42/54 =78%

## 5. Conclusions and Future

Our initial tests at a geological field site of this texture-based algorithm for image comparison show promise for both novelty detection and for similarity matching. The similarity matching was superb (91%), especially of images of the yellow lichens and of images of the coal bed. Our next step is for the Cyborg Astrobiologist to perform more extensive field testing of this algorithm with the Astrobiology phone-cam at other geological/astrobiological field sites.

Testing with gray-scale cameras and multispectral cameras are possible extensions of this work. To simulate images taken by a rover, one could take videos and randomly select frames from the video. We also hope to speed up the algorithm so that it can analyze ~100 images in a laptop-battery-limited real-time test of 1.5 hours. This speed-up can be accomplished perhaps by using the cluster characteristics of the image database so that each image is only compared to 1 image of each cluster of the database. The speed-up could also be accomplished by eliminating the delays of Bluetooth transmission and developing an app for smartphones which acts directly on the smartphone as soon as the image is captured; this would at least reduce the bottleneck to one battery life rather than two, and carrying a spare phone battery is much easier than carrying around a spare laptop battery. Another extension of the system could be useful for offline analysis: allowing for an image to be segmented for texture with this compression algorithm. This could lead the Cyborg Astrobiologist to understand further the images, so that the software could be used by a Robotic Astrobiologist to zero-in on novel astrobiological/geological areas of outcrops.

**Table 2** (Note: a new image database was created with the first image, which was image #264).

| Incoming Image # And Best matched pair | Matching Image # | Incoming Image Descrip. | Matching Image Descrip. | Mismatch Notes (Novel/Similar/Different *judged by human*) |
|---|---|---|---|---|
| **265** 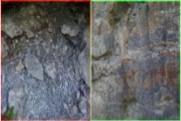 | **264** | loose weathered rock | thick-bedded sandstone with thin parting of mudstone in middle | **Novel.** **Only 2 images in database. $D_{sim}=0\%$** |
| **266** 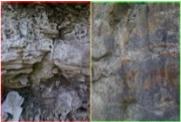 | **264** | thin-bedded sandstone | thick-bedded sandstone with thin parting of mudstone in middle | **Novel.** **Only 3 images in database. $D_{sim}=7\%$** |
| **267** 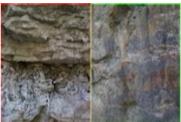 | **264** | thin-bedded sandstone below, shale above | thick-bedded sandstone with thin parting of mudstone in middle | **Novel.** **Only 4 images in database. $D_{sim}=11\%$** |
| **268** 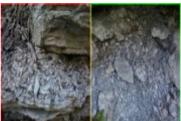 | **265** | weathered shale | loose weathered rock | **Novel.** **Only 5 images in database. $D_{sim}=25\%$** |
| **269** 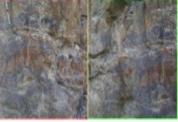 Name:Photo264.jpg Similarity = 39.5371% | **264** | thick-bedded sandstone with thin parting of mudstone in middle | thick-bedded sandstone with thin parting of mudstone in middle | **Similar.** $D_{sim}$=40% |
| **270** 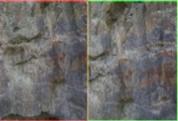 Name:Photo264.jpg Similarity = 40.7904% | **264** | thick-bedded sandstone with thin parting of mudstone in middle | thick-bedded sandstone with thin parting of mudstone in middle | **Similar.** $D_{sim}$=41% |
| **271** 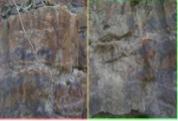 Name:Photo270.jpg Similarity = 39.4385% | **270** | thick-bedded sandstone with thin parting of mudstone in middle | thick-bedded sandstone with thin parting of mudstone in middle | **Similar.** $D_{sim}$=40% |
| **272** 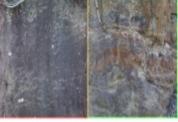 Name:Photo269.jpg Similarity = 44.0388% | **269** | massive sandstone | thick-bedded sandstone with thin parting of mudstone in middle | **Novel.** $D_{sim}$=44% **These 2 units appear very similar in the images.** |
| **273** 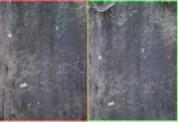 Name:Photo272.jpg Similarity = 48.1308% | **272** | massive sandstone | massive sandstone | **Similar.** $D_{sim}$=48% |
| **274** 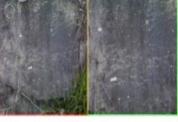 Name:Photo273.jpg Similarity = 44.5567% | **273** | massive sandstone | massive sandstone | **Similar.** $D_{sim}$=44% **Some grass in 274.** |
| **275** 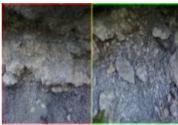 Name:Photo265.jpg Similarity = 41.5184% | **265** | nodular mudstone | loose weathered rock | **Novel.** $D_{sim}$=42% **These 2 units appear very similar in the images** |
| **276** 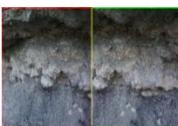 Name:Photo275.jpg Similarity = 46.1604% | **275** | nodular mudstone | nodular mudstone | **Similar.** $D_{sim}$=46% |
| **277** 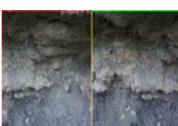 Name:Photo275.jpg Similarity = 47.5699% | **275** | nodular mudstone | nodular mudstone | **Similar.** $D_{sim}$=48% |
| **278** 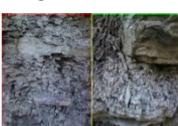 Name:Photo268.jpg Similarity = 25.7307% | **268** | laminated shale | weathered shale | **Novel.** $D_{sim}$=26% |
| **279** 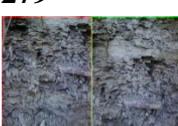 Name:Photo278.jpg Similarity = 37.4455% | **278** | laminated shale | laminated shale | **Similar.** $D_{sim}$=37% |
| **280** 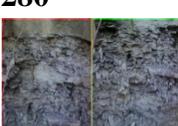 Name:Photo279.jpg Similarity = 37.2033% | **279** | laminated shale | laminated shale | **Similar.** $D_{sim}$=37% |
| **281** 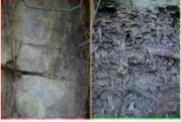 | **279** | laminated sandstone | laminated shale | **Novel.** $D_{sim}$=31% |
| **282** 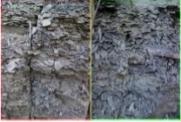 | **279** | laminated shale | laminated shale | **Similar.** $D_{sim}$=30% **Shale in 279 is much redder in color than in 282.** |
| **283** 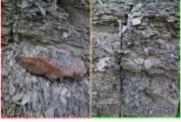 | **282** | laminated shale with ironstone concretion | laminated shale | **Novel.** $D_{sim}$=34% |
| **284** 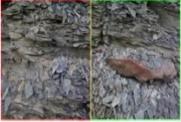 | **283** | laminated shale | laminated shale with ironstone concretion | **Different.** $D_{sim}$=31% **These two images are very similar. Ironstone is about 10% of image in area.** |
| **285** 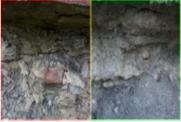 | **277** | laminated shale with ironstone concretion | nodular mudstone | **Different.** $D_{sim}$=41% **These two images are similar in color statistics** |
| **286** 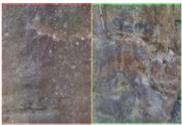 Name:Photo269.jpg Similarity = 46.538% | **269** | massive sandstone | thick-bedded sandstone with thin parting of mudstone in middle | **Different.** $D_{sim}$=46% These 2 units appear very similar in the images. |
| **287** 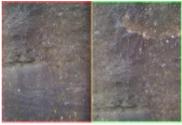 Name:Photo286.jpg Similarity = 49.0417% | **286** | massive sandstone | massive sandstone | **Similar.** $D_{sim}$=49% |
| **288** 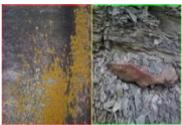 Name:Photo283.jpg Similarity = 29.8365% | **283** | sandstone with lichen | laminated shale w ironstone concretion | **Novel.** $D_{sim}$=30% |
| **289** 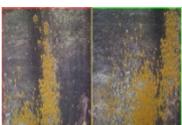 Name:Photo288.jpg Similarity = 40.9189% | **288** | sandstone with lichen | sandstone with lichen | **Similar.** $D_{sim}$=41% |
| **290** 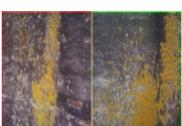 Name:Photo288.jpg Similarity = 39.9344% | **288** | sandstone with lichen | sandstone with lichen | **Similar.** $D_{sim}$=40% |
| **291** 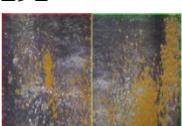 Name:Photo288.jpg Similarity = 43.006% | **288** | sandstone with lichen | sandstone with lichen | **Similar.** $D_{sim}$=43% |
| **292** 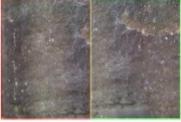 Name:Photo286.jpg Similarity = 49.2792% | **286** | massive sandstone | massive sandstone | **Similar.** **D$_{sim}$=49%** |
| **293** 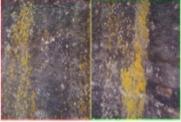 Name:Photo290.jpg Similarity = 43.7815% | **290** | sandstone with lichen | sandstone with lichen | **Similar.** **D$_{sim}$=44%** |
| **294** 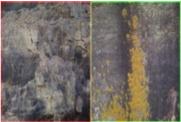 Name:Photo289.jpg Similarity = 43.3207% | **289** | Coal w cleats | sandstone with lichen | **Novel.** **D$_{sim}$=43%** **Sulfur coloring in color is also yellowish, like the lichens.** |
| **295** 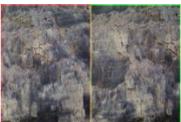 Name:Photo294.jpg Similarity = 44.7965% | **294** | Coal w cleats | Coal w cleats | **Similar.** **D$_{sim}$=45%** |
| **296** 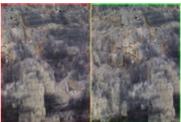 Name:Photo295.jpg Similarity = 47.131% | **295** | Coal w cleats | Coal w cleats | **Similar.** **D$_{sim}$=47%** |
| **297** 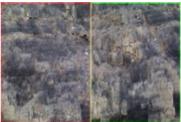 Name:Photo295.jpg Similarity = 41.7323% | **295** | Coal w cleats | Coal w cleats | **Similar.** **D$_{sim}$=42%** |
| **298** 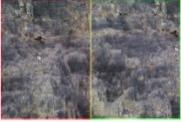 Name:Photo297.jpg Similarity = 42.9345% | **297** | Coal w cleats | Coal w cleats | **Similar.** **D$_{sim}$=43%** |
| **299** 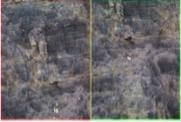 Name:Photo298.jpg Similarity = 41.4231% | **298** | Coal w cleats | Coal w cleats | **Similar.** **D$_{sim}$=41%** |
| **300** 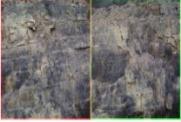 Name:Photo294.jpg Similarity = 45.1607% | **294** | Coal w cleats | Coal w cleats | **Similar.** **D$_{sim}$=45%** |
| **301** 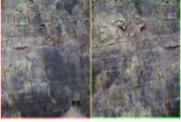 Name:Photo300.jpg Similarity = 46.6015% | **300** | Coal w cleats | Coal w cleats | **Similar.** **D$_{sim}$=47%** |
| **302** 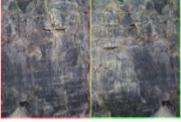 Name:Photo301.jpg Similarity = 49.4315% | **301** | Coal w cleats | Coal w cleats | **Similar.** **D$_{sim}$=49%** |
| **303** 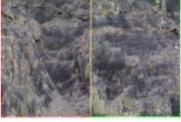 Name:Photo297.jpg Similarity = 43.2029% | **297** | Coal w cleats | Coal w cleats | **Similar.** **D$_{sim}$=43%** |
| **304** 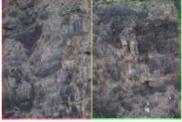 Name:Photo299.jpg Similarity = 45.3162% | **299** | Coal w cleats | Coal w cleats | **Similar.** $D_{sim}=45\%$ |
| **305** 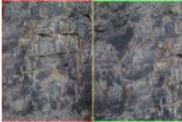 Name:Photo304.jpg Similarity = 45.2182% | **304** | Coal w cleats | Coal w cleats | **Similar.** $D_{sim}=45\%$ |
| **306** 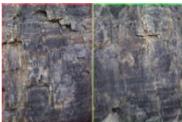 Name:Photo302.jpg Similarity = 45.734% | **302** | Coal w cleats | Coal w cleats | **Similar.** $D_{sim}=46\%$ |
| **307** 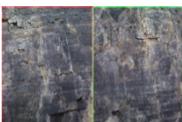 Name:Photo302.jpg Similarity = 49.0373% | **302** | Coal w cleats | Coal w cleats | **Similar.** $D_{sim}=49\%$ |
| **308** 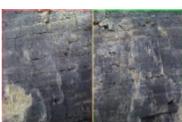 Name:Photo307.jpg Similarity = 45.486% | **307** | Coal w cleats | Coal w cleats | **Similar.** $D_{sim}=45\%$ |
| **309** 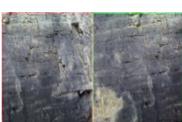 Name:Photo308.jpg Similarity = 50.2931% | **308** | Coal w cleats | Coal w cleats | **Similar.** $D_{sim}=50\%$ |

| 310 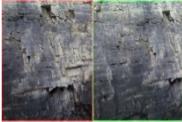 | 309 | Coal w cleats | Coal w cleats | **Similar.** **D$_{sim}$=48%** |
|---|---|---|---|---|
| 311 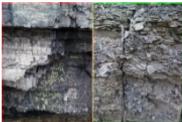 | 282 | Coal w yellow growths | laminated shale | **Novel.** **D$_{sim}$=41%** **These images are similar to each other in texture and color statistics.** |
| 312 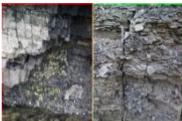 | 282 | Coal w yellow growths | laminated shale | **Different.** **D$_{sim}$=38%** **These images are similar to each other in color statistics.** |
| 313 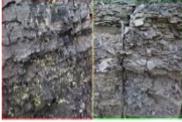 | 282 | Coal w yellow growths | laminated shale | **Different.** **D$_{sim}$=41%** **These images are similar to each other in texture and color statistics.** |
| 314 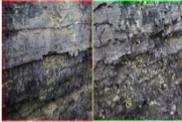 | 313 | Coal w yellow growths | Coal w yellow growths | **Similar.** **D$_{sim}$=47%** |

| 315 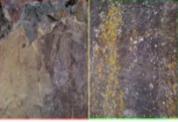 | 293 | limestone | sandstone with lichen | **Novel.** $D_{sim}$=**48%** **These images are similar to each other in color statistics.** |
| --- | --- | --- | --- | --- |
| 316 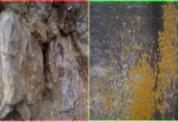 | 288 | limestone with cracks | sandstone with lichen | **Novel.** $D_{sim}$=**39%** **These images are similar to each other in color statistics.** |
| 317 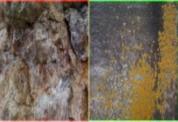 | 288 | limestone with cracks | sandstone with lichen | **Different.** $D_{sim}$=**43%** **These images are similar to each other in color statistics.** |
| 318 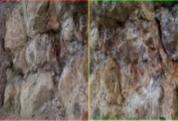 | 317 | limestone with cracks | limestone with cracks | **Similar.** $D_{sim}$=**42%** |

**Acknowledgements**

We thank Helge Ritter for suggesting the image-compression technique to us, and Tim Warner for help with identifying a field site in West Virginia. PCM acknowledges past support by a research fellowship from the Alexander von Humboldt Foundation, support from the Freie Universität Berlin, and computer resources for the preparation of this manuscript at the library of West Virginia University. CG acknowledges the Helmholtz Association through the research alliance "Planetary Evolution and Life". The work by JO is supported by grants AYA2008-03467/ESP and AYA2011-24780/ESP from the Spanish Ministry of Economy and Competitiveness.

**References**

Bartolo, A., P.C. McGuire, K.P. Camilleri, C. Spiteri, J.C. Borg, P.J. Farrugia, J.Ormö, J. Gomez Elvira, J.A. Rodriguez Manfredi, E. Diaz Martinez, H. Ritter, R. Haschke, M. Oesker, J. Ontrup (2007), "The Cyborg Astrobiologist: Porting from a Wearable Computer to the Astrobiology Phone-cam", *International Journal of Astrobiology*, **6**, pp. 255-261.

Bonnici, A., C. Gross, P.C. McGuire, J. Ormö, S. Walter, and L. Wendt (2010), "The Cyborg Astrobiologist: Compressing Images for the Matching of Prior Textures and for the Detection of Novel Textures**",** *European Planetary Science Congress (EPSC)*, Potsdam, Germany, **5**, extended abstract #162.

Castano, R., T. Estlin, D. Gaines, C. Chouinard, B. Bomstein, R.C. Anderson, M. Burl, D. Thompson, A. Castano, and M. Judd (2007), "Onboard Autonomous Rover Science", *IEEE Aerospace Conf.,* Big Sky, Montana.

Fink, W., J.M. Dohm, M.A. Tarbell, T.M. Hare and V.R. Baker (2005), "Next-generation robotic planetary reconnaissance missions: A paradigm shift", *Planetary and Space Science*, **53**, pp. 1419-1426.

Gross, C., L. Wendt, P.C. McGuire, A. Bonnici, B.H. Foing, V. Souza-Egipsy, R. Bose, S. Walter, J. Ormö, E. Diaz-Martinez, M. Oesker, J. Ontrup, R. Haschke, H. Ritter (2010), "The Cyborg Astrobiologist: testing a novelty detection algorithm at the Mars Desert Research Station (MDRS), Utah," *LPSCXLI, Lunar and Planetary Science Conf.*, Houston, Texas, extended abstract #2457.

Gross, C., L. Wendt, P.C. McGuire, A. Bonnici, B.H. Foing, V. Souza-Egipsy, R. Bose, S. Walter, J. Ormö. E. Diaz-Martinez, M. Oesker, J. Ontrup, R. Haschke, H. Ritter (2009), "Testing the Cyborg Astrobiologist at the Mars Desert Research Station (MDRS),

Utah", *European Planetary Science Congress (EPSC)*, Potsdam, Germany, **4**, extended abstract #548.

Gulick, V.C., Morris, R. L., Ruzon, M. A. and Roush, T. L. (2001), "Autonomous image analyses during the 1999 Marsokhod rover field test", *Journal of Geophysical Research,* **106**, pp. 7745-7764.

Gulick, V.C., Hart, S. D., Shi, X. and Siegel, V. L. (2004). "Developing an automated science analysis system for Mars surface exploration for MSL and beyond". *Lunar and Planetary Institute Conference XXXV*, extended abstract #2121.

Halatci, I., C.A. Brooks and K. Iagnemma (2008), "A study of visual and tactile terrain classification and classifier fusion for planetary exploration rovers". *Robotica,* **26**, pp. 767-779.

Halatci, I, C.A. Brooks, and K. Iagnemma (2007), "Terrain classification and classifier fusion for planetary exploration rovers", *IEEE Aerospace Conference*, Big Sky, MT, pp. 1-11.

Haralick, R.M., K. Shanmugam, and I.H. Dinstein (1973), "Textural features for image classification", *IEEE Transactions on Systems, Man and Cybernetics,* **6**, pp. 610-621.

Heidemann, G. and H. Ritter (2008a), "Compression for Visual Pattern Recognition", *Proceedings of the Third International Symposium on Communications, Control and Signal Processing (ISCCSP 2008)*, St. Julians, Malta, IEEE, pp. 1520-1523.

Heidemann, G., and H. Ritter (2008b), "On the contribution of compression to visual pattern recognition", In *Proc. 3rd Int'l Conf. on Comp. Vision Theory and Applications, Funchal, Madeira-Portugal*, **2**, pp. 83-89.

Huffman, D.A. (1952), "A Method for the Construction of Minimum-Redundancy Codes", *Proceedings of the Institute of Radio Engineers (IRE)*, pp. 1098–1102.

Kärkkäinen, J., D. Kempa, and S.J. Puglisi (2013), "Lightweight Lempel-Ziv Parsing", *Experimental Algorithms*, Springer Berlin Heidelberg, pp. 139-150.
http://arxiv.org/abs/1302.1064

Lempel, A. and J. Ziv (1977), "A Universal Algorithm for Sequential Data Compression", *IEEE Trans. Inf. Th.*, **23**, pp. 337–343.

McGuire, P.C., J.O. Ormö, E. Diaz Martinez, J.A. Rodriguez Manfredi, J. Gomez Elvira, H. Ritter, M. Oesker, J. Ontrup (2004), "The Cyborg Astrobiologist: First Field Experience", *International Journal of Astrobiology*, **3**, pp. 189-207.

McGuire, P.C., E. Diaz Martinez, J.O. Ormö, J. Gomez Elvira, J.A. Rodriguez Manfredi, E. Sebastian Martinez, H. Ritter, R. Haschke, M. Oesker, J. Ontrup (2005a), "The Cyborg Astrobiologist: Scouting Red Beds for Uncommon Features with Geological Significance", *International Journal of Astrobiology*, **4**, pp. 101-113.

McGuire, P.C., J. Gomez Elvira, J.A. Rodriguez Manfredi, E. Sebastian Martinez, J. Ormö, E. Diaz Martinez, H. Ritter, M. Oesker, R. Haschke and J. Ontrup (2005b), "Field Geology with a Wearable Computer: First Results of the Cyborg Astrobiologist System", *Proceedings of the ICINCO'2005 (International Conference on Informatics in Control, Automation and Robotics)*, September 14-17, Barcelona, Spain, **3**, pp. 283-291.

McGuire, P.C., M.J. Wolff, M.D. Smith, R.E. Arvidson, S.L. Murchie, R.T. Clancy, T.L. Roush, S.C. Cull, K.A. Lichtenberg, S.M. Wiseman, R.O. Green, T.Z. Martin, R.E. Milliken, P.J. Cavender, D.C. Humm, F.P. Seelos, K.D. Seelos, H.W. Taylor, B.L. Ehlmann, J.F. Mustard, S.M. Pelkey, T.N.Titus, C.D. Hash, E.R. Malaret, and the CRISM Team (2008), "MRO/CRISM Retrieval of Surface Lambert Albedos for Multispectral Mapping of Mars with DISORT-based Radiative Transfer Modeling: Phase 1 -- Using Historical Climatology for Temperatures, Aerosol Optical Depths, and Atmospheric Pressures", *Transactions on Geoscience and Remote Sensing*, **46**, pp. 4020-4040.

McGuire, P.C., C. Gross, L. Wendt, A. Bonnici, V. Souza-Egipsy, J. Ormö, E. Diaz-Martinez, B.H. Foing, R. Bose, S. Walter, M. Oesker, J. Ontrup, R. Haschke, H. Ritter (2010), "The Cyborg Astrobiologist: Testing a Novelty-Detection Algorithm on Two Mobile Exploration Systems at Rivas Vaciamadrid in Spain and at the Mars Desert Research Station in Utah", *International Journal of Astrobiology*, **9**, pp. 11-27.

Rao, A.R. (2012), *A taxonomy for texture description and identification*. Springer Publishing Company, Incorporated.

Saravanan, C., and R. Ponalagusamy (2010), "Lossless Grey-scale Image Compression using Source Symbols Reduction and Huffman Coding", *International Journal of Image Processing (IJIP),* **3**, pp. 246-251.

Thompson, D.R., A. Allwood, D. Bekker, N.A. Cabrol, T. Estlin, T. Fuchs, K.L.Wagstaff (2012), "TextureCam: Autonomous Image Analysis for Astrobiology Survey", *LPSCXLIII Lunar and Planetary Institute Conference,* Houston, Texas, extended abstract #1659.

Volpe, R. (2003). "Rover functional autonomy development for the Mars Mobile Science Laboratory". *IEEE Aerospace Conference*, Big Sky, Montana. **2**, pp. 2_643-2_652.

Wendt, L., C. Gross, P.C. McGuire, A. Bonnici, B.H. Foing, V. Souza-Egipsy, R. Bose, S. Walter, J. Ormö. E. Diaz-Martinez, M. Oesker, J. Ontrup, R. Haschke, H. Ritter (2009), "The Cyborg Astrobiologist: Teaching Computers to Find Uncommon or Novel Areas of Geological Scenery in Real-time", *European Space Agency International Conference on Comparative Planetology: Venus - Earth - Mars*, ESTEC, Noordwijk, The Netherlands.